\documentclass[conference]{QC-Explaining-Generalization}
\usepackage{cite} 
\usepackage{amsmath,amssymb,amsfonts} 
\usepackage{algorithmic} 
\usepackage{textcomp} 
\usepackage{xcolor} 
\usepackage{hyperref}
\usepackage{url}
\usepackage{graphicx} 
\graphicspath{ {./images/} }
\usepackage{float} 
\usepackage{multicol}
\usepackage{multirow}
\usepackage{xcolor}
\usepackage{subfigure}
\usepackage{caption}

\makeatletter
\def\hlinewd#1{%
  \noalign{\ifnum0=`}\fi\hrule \@height #1 \futurelet
  \reserved@a\@xhline}

\def\BibTeX{{\rm B\kern-.05em{\sc i\kern-.025em b}\kern-.08em 
    T\kern-.1667em\lower.7ex\hbox{E}\kern-.125emX}}

\begin{document}
\title{In Search of Probeable Generalization Measures\\}

\author{\IEEEauthorblockN{Jonathan Jaegerman{$^1$}\thanks{$^*$ Equal Contribution}$^*$,
Khalil Damouni{$^1$}\IEEEauthorrefmark{1}
Mahdi S. Hosseini{$^2$}$^*$,
Konstantinos N. Plataniotis{$^1$}
}
\IEEEauthorblockA{{$^1$}The Edward S. Rogers Sr. Department of Electrical \& Computer Engineering, University of Toronto}
\IEEEauthorblockA{{$^2$}The Department of Electrical and Computer Engineering, University of New Brunswick}
\IEEEauthorblockA{\url{https://github.com/mahdihosseini/GenProb}}
}

\maketitle

\begin{abstract}
Understanding the generalization behaviour of deep neural networks is a topic of recent interest that has driven the production of many studies, notably the development and evaluation of generalization ``explainability'' measures that quantify model generalization ability. Generalization measures have also proven useful in the development of powerful layer-wise model tuning and optimization algorithms, though these algorithms require specific kinds of generalization measures which can probe individual layers. The purpose of this paper is to explore the neglected subtopic of probeable generalization measures; to establish firm ground for further investigations, and to inspire and guide the development of novel model tuning and optimization algorithms. We evaluate and compare measures, demonstrating effectiveness and robustness across model variations, dataset complexities, training hyperparameters, and training stages. We also introduce a new dataset of trained models and performance metrics, GenProb, for testing generalization measures, model tuning algorithms and optimization algorithms.
\end{abstract}


\section{Introduction}
Deep learning has proven very successful this last decade, demonstrating time and time again its ability to generalize feature recognition from train data to test data. Despite all the attention, the underlying mechanisms in deep models that promote generalization are still open questions \cite{huh2021lowrank}. A number of papers attempt to consolidate the intuition behind deep learning generalization by developing "explainable" measures that attempt to quantify the generalization ability of a given model and dataset \cite{jiang2019fantastic, dziugaite2021search, neyshabur2017exploring}.

Generalization measures provide understanding of the learning mechanisms involved in the training of a given network using certain optimizers, datasets, and hyperparameters. Understanding and identifying trends in the relationship between these metrics and the network’s generalization gap or test accuracy provides methods of optimally selecting hyperparameter configuration, topologies, and other training parameters. Furthermore, generalization measures can be implemented for neural architecture search (NAS), hyperparameter optimization (HPO) and training optimization \cite{NIPS2015_eaa32c96, DBLP:journals/corr/abs-2010-01412, cherian2020efficient,DBLP:journals/corr/abs-1802-07191, abdelfattah2021zerocost}. Other methods optimize deep models solely via holistic evaluations of performance, neglecting the discrepancies in training quality between layers, and foregoing the additional tuning required for a better solution.

\begin{figure}[t]
    \centering
    \includegraphics[width=0.5\textwidth]{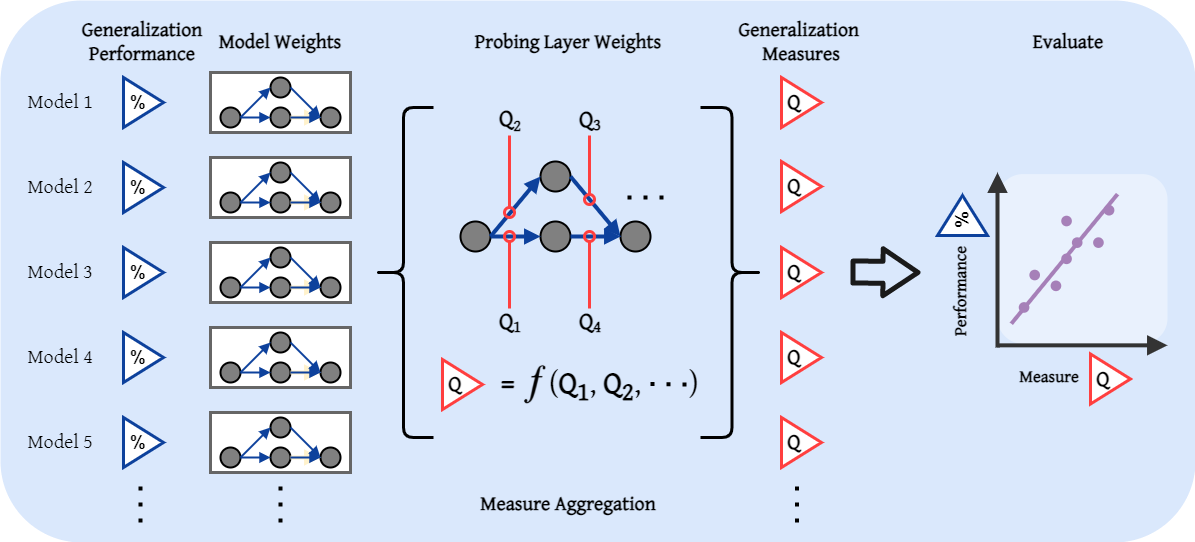}
    \caption{Probeable Generalization Measure Evaluation Pipeline}
    \label{genevalpipe}
\end{figure}

Many generalization measures have proven effective and robust, but are not practical for implementation of model tuning and optimization algorithms. It is optimal that a generalization measure can probe individual layers (a probeable measure), in order to tune models at a layer level (e.g. channel sizes, weight update rules, individual learning rates). The most successful measures are implemented from PAC-Bayes bounds and margin distributions, and are not probeable \cite{jiang2019fantastic, dziugaite2021search, natekar,k2021robustness}. Our objective in this paper is to investigate probeable generalization measures for model tuning and optimization applications, and to better understand generalization mechanisms in deep models. Given a dataset of trained models and their performance metrics, we evaluate probeable generalization metrics by the pipeline depicted in \autoref{genevalpipe}.

The following lists the contributions in this paper:
 \begin{itemize}
     \item We describe and investigate four probeable generalization measures that can be directly measured from individual layers of a deep network, without the need for additional training pipeline.
     \item We test these measures on the NATS-Bench dataset \cite{Dong_2021}, and demonstrate their effectiveness for explaining the generalization of models of varying channel sizes.
     \item We study the evolution of these measures during training to understand their effectiveness at different stages of training.
     \item We introduce a new dataset--dubbed GenProb--to test probeable measures on models of varying channel sizes and training hyperparameters.
     \item We evaluate and compare the measures with GenProb, demonstrating effectiveness and robustness.
 \end{itemize}
\section{Related Works}
With a recent surge in interest for understanding the generalization performance of deep models, studies have produced many new complexity measures. Response to input perturbation measures \cite{DBLP:journals/corr/abs-2106-04765,natekar,k2021robustness}, and PAC-Bayes sharpness and flatness measures \cite{natekar,DBLP:journals/corr/DziugaiteR17,jiang2019fantastic} have demonstrated much success. Notably, the first and second place holders \cite{natekar,k2021robustness} of the recent NeurIPS 2020: Predicting Generalization in Deep Learning competition \cite{jiang2020neurips}, developed perturbation and sharpness measures. These measures have also been implemented in NAS, HPO and training optimization algorithms \cite{DBLP:journals/corr/abs-2010-01412, cherian2020efficient,DBLP:journals/corr/abs-1802-07191}.

Studies also define measures from deep network margin distributions \cite{DBLP:journals/corr/abs-2106-03314,arora2018stronger,bartlett2017spectrallynormalized}. Some other measures are derived from model weights themselves, such as gradient signal-to-noise-ratio (GSNR) \cite{liu2020understanding} and norm-based measures \cite{NIPS2015_eaa32c96,DBLP:journals/corr/abs-1901-01672,DBLP:journals/corr/abs-1711-01530,novak2018sensitivity}. Shallow networks are often trained onto the aforementioned measures for accurate generalization performance predictions, at the cost of ability to generalize predictions across different datasets and model structures \cite{jiang2018predicting,yak2019task,corneanu2020computing,unterthiner2021predicting}.
Some large scale studies of generalization measures provide vigorous evaluations and comparisons of common and best performing measures, across different datasets and model structures \cite{jiang2019fantastic, dziugaite2021search, neyshabur2017exploring}. 

These last three are the only studies that include probeable measures, however they are tested on sets of fully trained models with drastic variations; irrelevant for most model optimization algorithms. Probeable measures are implemented in \cite{abdelfattah2021zerocost} to reduce the computational cost and increase performance of NAS algorithms, and in \cite{DBLP:journals/corr/abs-2006-06587} to adaptively optimize deep networks for greater generalization at no additional computational overhead.

\section{Metric Definitions}
We define several measures that quantify the quality of a trained model (i.e. quality metrics or complexity measures) and describe its generalization ability. These quality metrics are probeable on individual layers of deep network, and quantify the contribution of each layer as a holistic measure for network representation, unlike other popular and successful measures elaborated above. The measures are thus explainable as they indicate how well training is optimized across layers of a deep network. The overview of all chosen measures is presented in \autoref{table_metric_defnition}.
\begin{table}[b]
\caption{Quality Metric Definitions}
\label{table_metric_defnition}
\begin{center}
\renewcommand{\tabcolsep}{1pt}
\scriptsize{
\begin{tabular}{c||c||c}
\hlinewd{1pt}
    \textbf{Metric} & \textbf{Formulation} & \textbf{Aggregation} \\
    \hline\hline
    Stable Quality 
    &\( 
    arctan\; s(\boldsymbol{W}_i) / \kappa(\boldsymbol{W}_i)\) 
    & \(
    \prod_{i=1}^d (Q_{SQ}(\boldsymbol{W}_i))^{1/d}\)\\ 
    \hline
    Effective Rank
    &\(
    \sum_{i=1}^{n^{\prime}}{\overline{\sigma}}_k(\boldsymbol{W}_i)log({\overline{\sigma}}_k(\boldsymbol{W}_i))\) & \(
    log(\sqrt{\sum_{i=1}^d {Q_{ER}(\boldsymbol{W}_i)}^2 / d})\)\\
    \hline
    Spectral Norm 
    & \(
    max(\boldsymbol{\sigma}(\boldsymbol{W}_l))\)& \(
    log\;\sqrt{d(\prod_{i=1}^d \|\boldsymbol{W}_i\|_2^2)^{1/d}}\) \\
    \hline
    Frobenius Norm
    & \(
    \sqrt{\sum_{j=1}^{m}\sum_{k=1}^{n}|w_{ijk}|^2}\)&  \(
    log\;\sqrt{d(\prod_{i=1}^d \|\boldsymbol{W}_i\|_F^2)^{1/d}} \)  \\
    \hlinewd{1pt}
\end{tabular}
}
\end{center}
Here $d$ is the depth of the model, $n$ is the maximum rank of the weight matrix, $n^{\prime}$ is the rank of the weight matrix, $\boldsymbol{\sigma}(\boldsymbol{W}_i)$ is the vector of singular values of matrix $\boldsymbol{W}_i$ with individual values $\sigma_k(\boldsymbol{W}_i)$ and normalized singular values $\overline{\sigma}_k(\boldsymbol{W}_i)$ and $w_{ijk}$ is the value at row $j$ and column $k$ of matrix $\boldsymbol{W}_i$. Four dimensional weight tensors (such as convolutions) are first unfolded along their input and output channels for computation. 
\end{table}

\textit{Stable quality (SQ)} refers to the stability of encoding in a deep layer that is calculated with the relative ratio of stable rank and condition number of a layer defined in \cite{DBLP:journals/corr/abs-2006-06587}. Stable rank encodes the space expansion under the matrix mapping of the layer, and condition number indicates the numerical sensitivity of the mapping layer. Altogether the measure introduces a quality measure of the layer as an autoencoder.

\textit{Effective rank (E)} refers to the dimension of the output space of the transformation operated by a deep layer that is calculated with the Shannon entropy of the normalized singular values of a layer as defined in \cite{royoliv}.

\textit{Frobenius norm (F)} refers to the magnitude of a deep layer that is calculated with the sum of the squared values of a weight tensor. Frobenius norm is also calculated with the sum of the squared singular values of a layer.

\textit{Spectral norm (S)} refers to the maximum magnitude of mapping by a transformation operated by a layer that is calculated as the maximum singular value of a weight tensor.

The formulations of these quality metrics are presented in \autoref{table_metric_defnition}. With these layer-level quality measures, we aim to aggregate across model layers for a single meaningful model-level quality metric. For the norm-based measures we borrow the best performing aggregation method for each layer-level measure from \cite{dziugaite2021search}. For stable quality and effective rank, we inspire ourselves from literature aggregation methods, test all variations, and choose a single best performing method \cite{dziugaite2021search, jiang2019fantastic}.

The notation convention used in \autoref{table_metric_defnition} and hereinafter to represent different quality metrics is: $Q^{AGG}_M$ where aggregation $AGG\in\{L2=$ depth-normalized L2 norm, $p=$ depth-normalized product$\}$ and metric $M\in\{SQ =$ stable quality, $E =$ effective rank, $F =$ Frobenius norm, $S =$ stable norm$\}$. 

Low-rank factorization (LRF) is a preprocessing technique we also employ, aiming to increase the consistency of quality metrics across stages of training. By LRF, the low-rank component of a weight matrix, which represents the useful information, is extracted from raw weights, stripping the weights of residual noise from random initialization. We compute the factorization of our weight matrices by means of EVBMF defined in \cite{NIPS2011_b73ce398}. A wide hat~$\widehat{}$~can be included in the notation as $\widehat{Q}^{AGG}_M$ to indicate preprocessing of weights by low rank factorization.

\section{Preliminary Experiments}
\subsection{Quality Metrics and Varying Network Architecture}
NATS-Bench is a dataset of trained deep neural networks for benchmarking neural architecture search algorithms \cite{Dong_2021}. A set of various related model architectures are trained and have their weights and performances saved to generate NATS-Bench, a mapping of model architectures and weights to generalization performance. NATS-Bench models are designed with a standard architecture skeleton as described in \cite{Dong_2021}, and vary layer operations or channel sizes, at different locations in the skeleton. We can test the quality metrics with NATS-Bench by computing them on the provided trained weights, and correlating them with model generalization gap and test accuracy.
\begin{table}[b]
\caption{NATS-Bench Hyperparameter Variation Summary}
\label{NATSbenchhp}
\begin{center}
\renewcommand{\arraystretch}{1}
\scriptsize{
\begin{tabular}{c||c|c}
    \hlinewd{1pt}
    \textbf{Hyperparameter} & \textbf{Size Search Space} & \textbf{Topology Search Space}\\ \hline\hline
    Learning Rate & \multicolumn{2}{c}{\(0.1\rightarrow 0\) (cosine)} \\\hline
    Weight Decay  & \multicolumn{2}{c}{5e-4} \\\hline
    Batch Size    & \multicolumn{2}{c}{256}               \\\hline
    Epochs        & 12, 90 & 12, 200             \\\hline
    \multirow{2}{*}{Channel Size Variations} & 8, 16, 24, 32, & \multirow{2}{*}{-}\\
    & 40, 48, 56, 64 & \\\hline
    \multirow{2}{*}{Layer Operation Variations} & \multirow{2}{*}{-} & zeroize, skip, 1x1 conv, \\
    & & 3x3 conv, average-pool\\
    \hlinewd{1pt}
\end{tabular}
}
\end{center}
\end{table}

The NATS-Bench dataset consists of two subsets: the topology search space contains data from a set of models of varying layer operations, and the size search space contains data from a set of models of varying layer channel sizes. The topology search space features a set of five operations (\autoref{NATSbenchhp}) for six layers (15,625 total permutations), and the size search space features a set of eight channel sizes (\autoref{NATSbenchhp}) for five layers (32,768 total permutations). The models are trained on CIFAR10, CIFAR100 and ImageNet-16-120 datasets \cite{chrabaszcz2017downsampled}. Model weights were saved at 12 epochs, and at completion. Training was optimized with Nesterov momentum SGD for cross-entropy loss, with L2 weight decay and cosine annealing (see \autoref{NATSbenchhp} for hyperparameter selection). Data augmentation included random flips with probability of 0.5, 32x32 (16x16 for ImageNet-16-120) random crops with 4 pixel padding, and RGB channel normalization.

\begin{figure}[!b]
\vspace{-2mm}
    \centering
    \includegraphics[width=0.2\textwidth]{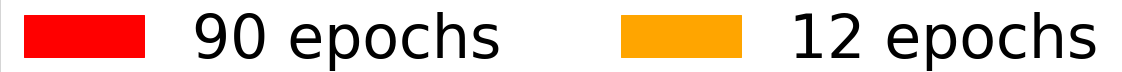}
    \subfigure[Size Search Space]{
    \includegraphics[width=0.23\textwidth]{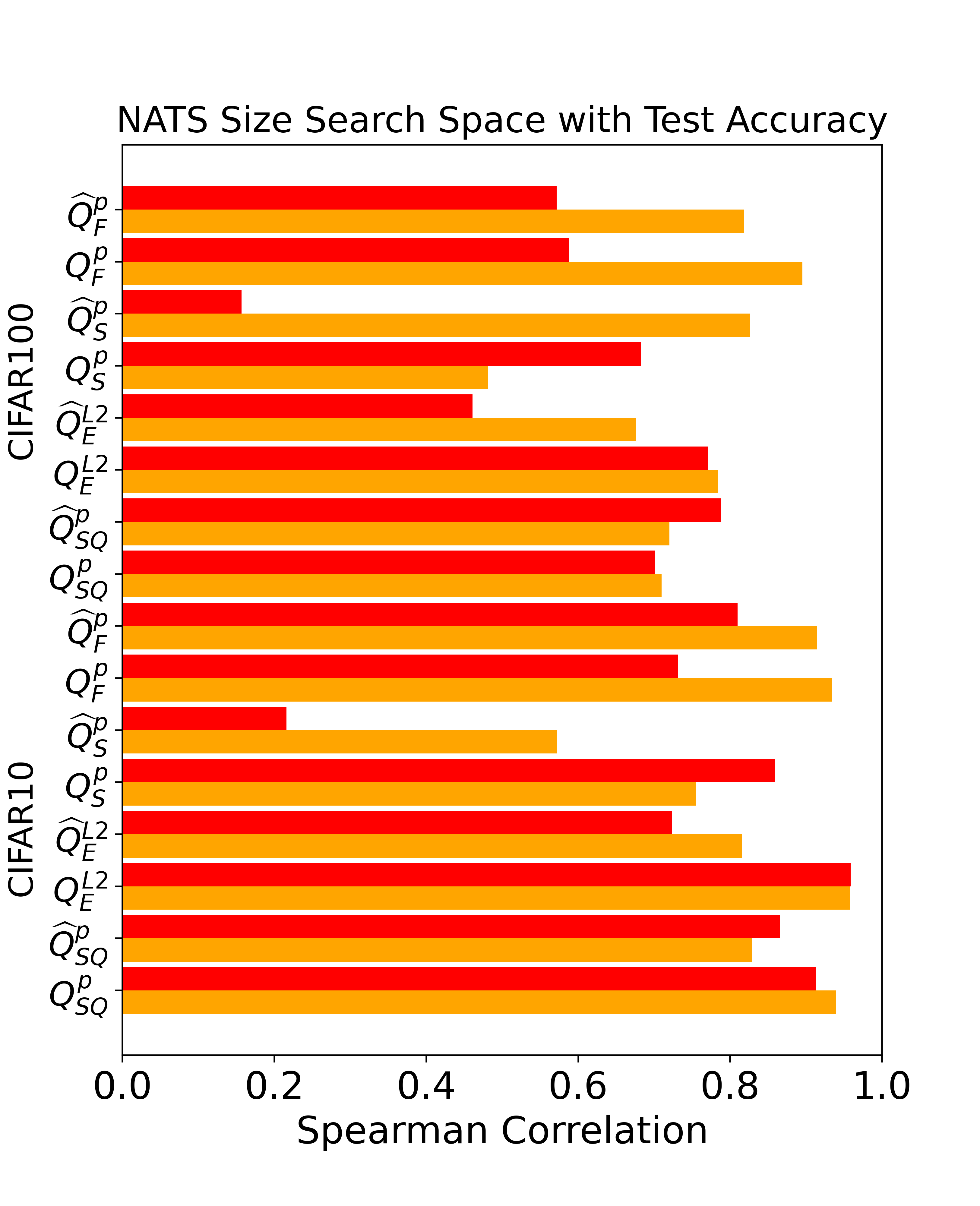}
    \includegraphics[width=0.23\textwidth]{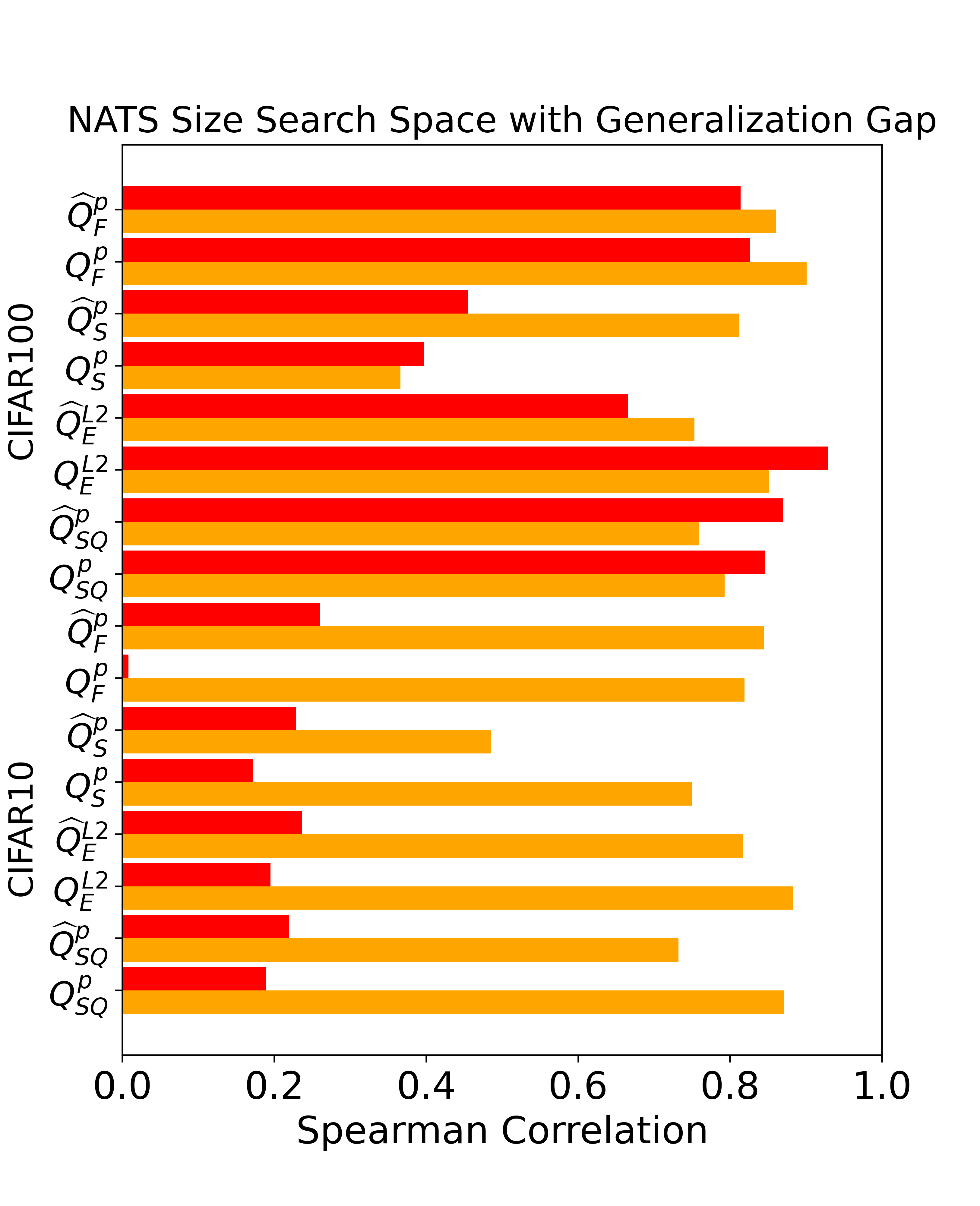}
    \label{NATSS}
    }
    \includegraphics[width=0.2\textwidth]{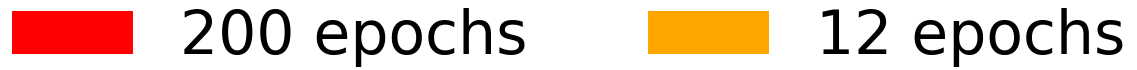}
    \subfigure[Topology Search Space]{
    \includegraphics[width=0.23\textwidth]{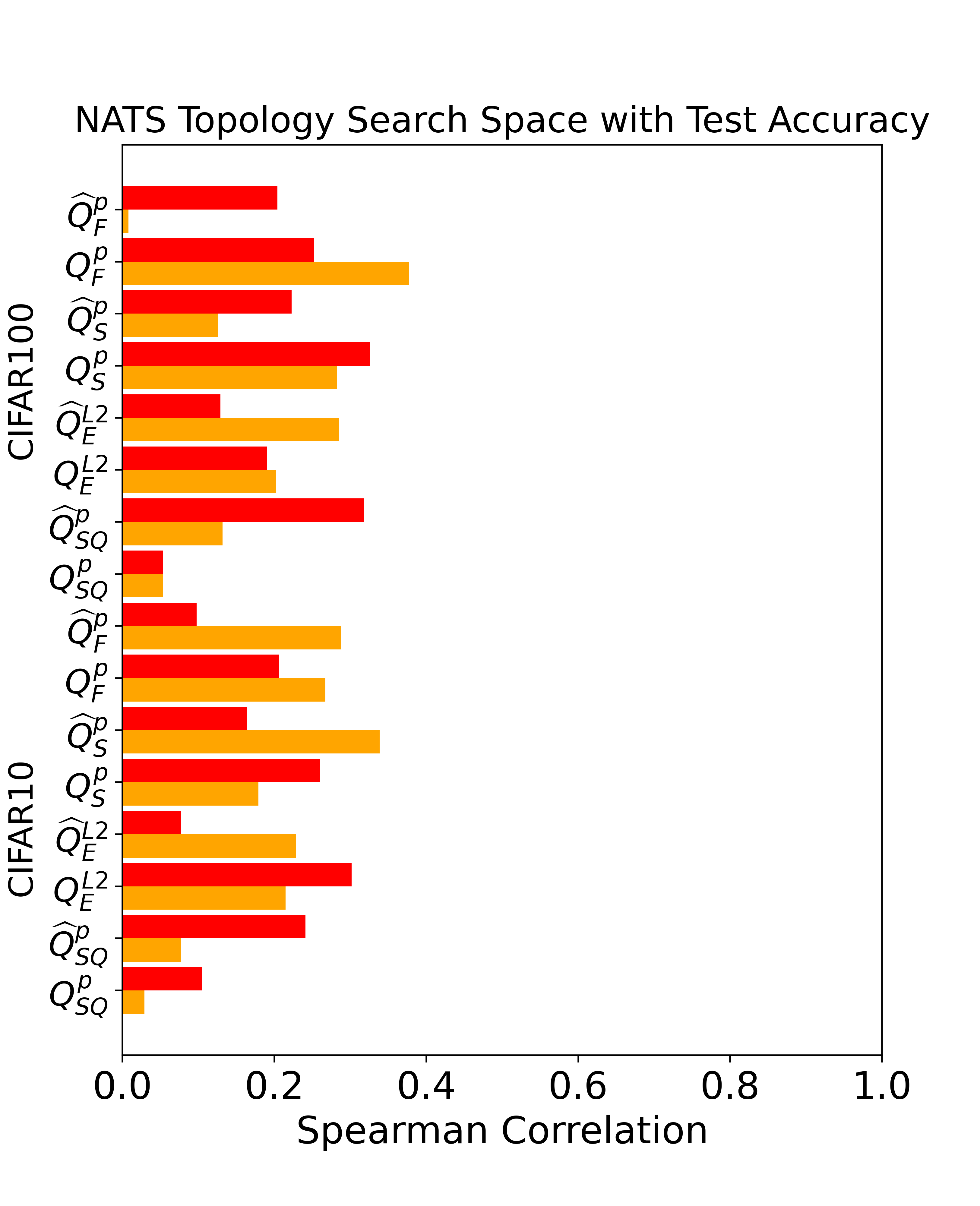}
    \includegraphics[width=0.23\textwidth]{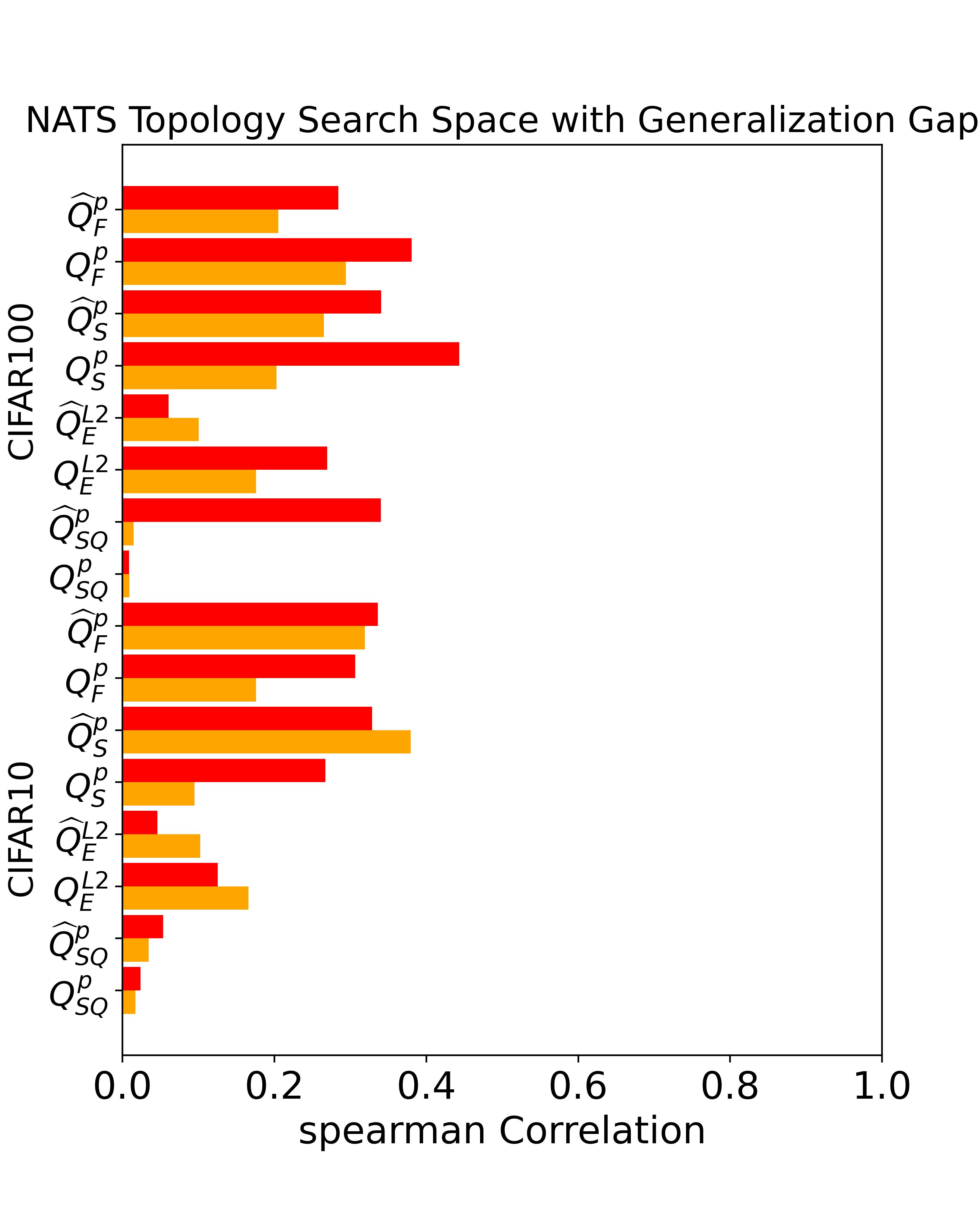}
    \label{NATST}
    }
    \caption{Evolution of Generalization Measures during Training with and without LRF}
    \label{metreval}
\end{figure}

Quality metrics are computed with and without use of the Low-Rank Factorization (LRF) on each set of trained model weights, then the Spearman correlations of these measures, with test accuracy and generalization gap, are computed to generate \autoref{metreval}. Spearman rank-order correlation describes the quality of a relationship between two variables as an arbitrary non-parametric monotonic function. Spearman correlation therefore fits our purpose of evaluating generalization measures, as a generalization measure only needs to rank the relative performance of competing models.

Several measures show very high correlations with test accuracy and generalization gap in the NATS size search space, reaching to and above 0.9, as shown in \autoref{NATSS}. Most raw quality metrics show a lot of potential, though use of LRF preprocessing of weights does not provide a consistent improvement. A trend is not observed across different epochs; the patterns for 12 epochs can’t be found in the 90 and 200 epoch alternative, indicating measure inconsistency across stages of training. Trends only hold across datasets in the case of NATS size search space for test accuracy, also indicating measure inconsistency across datasets. The patterns from the size search space do not appear for the topology search space; as illustrated in \autoref{NATST}, correlations don't pass 0.5, and many of the previous peaks are now troughs. The measures are not valuable across models of changing topologies, as they are sensitive to and vary with the types of layers included in the model.

The NATS-Bench search spaces feature drastic changes between models which yield stochastic results and inhibit meaningful analysis; however, the high correlations under the size search space motivate further investigation. Notably for HPO and training optimization algorithms, it would be valuable to use a search space with variations in training hyperparameters instead. A last remark is that LRF boosts correlations for certain configurations (e.g. stable quality in the topology search space), which warrants further investigation.

\subsection{Evolution of Quality Metrics during Training}
We trained ResNet34 on CIFAR10 and CIFAR100 with AdaS \cite{DBLP:journals/corr/abs-2006-06587} until completion and saved model weights at every epoch. The saved set of weights enables the observation of the evolution of quality metrics over training.

\begin{figure}[!t]
\vspace{-2mm}
    \centering
    \includegraphics[width=0.22\textwidth]{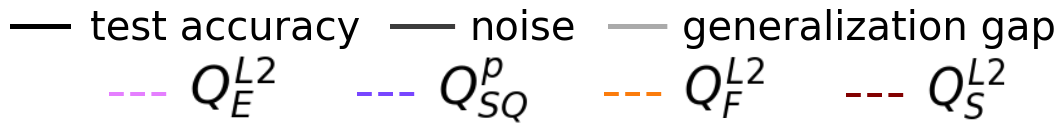}\hspace*{5pt}
    \includegraphics[width=0.22\textwidth]{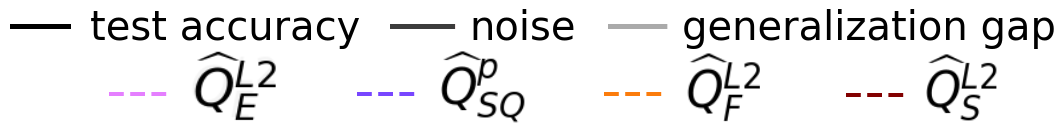}
    \subfigure[CIFAR-10 Results]{
    \includegraphics[width=0.23\textwidth]{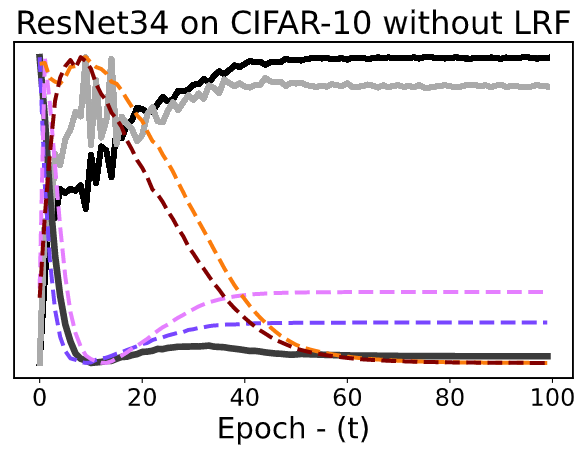}
    \includegraphics[width=0.23\textwidth]{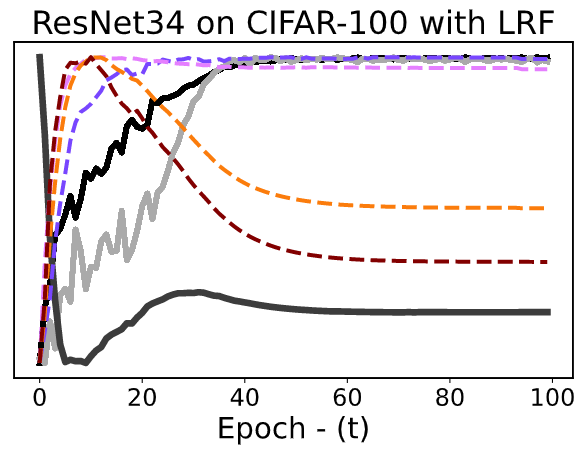}
    \label{lrfcifar10}
    }
    \subfigure[CIFAR-100 Results]{
    \includegraphics[width=0.23\textwidth]{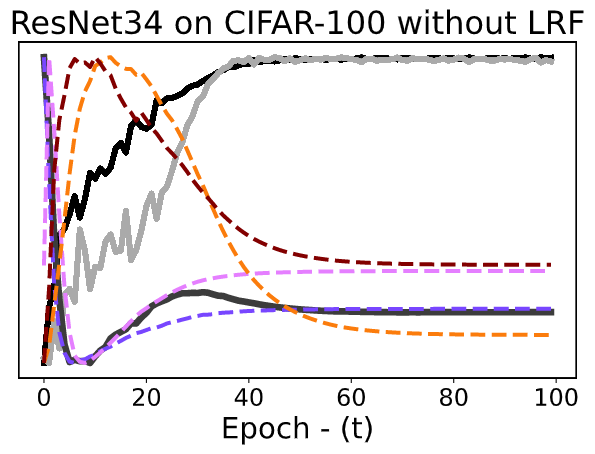}
    \includegraphics[width=0.23\textwidth]{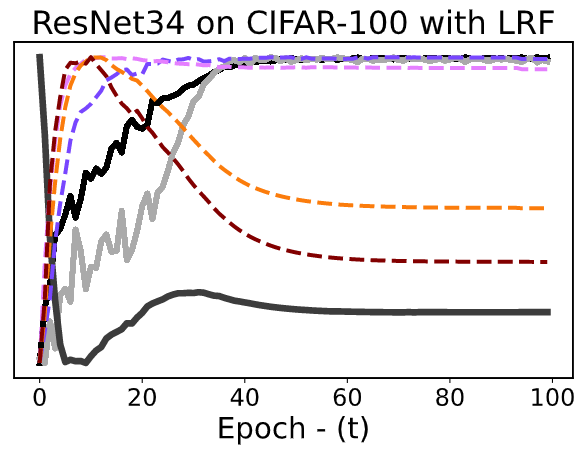}
    \label{lrfcifar100}
    }
    \caption{Evolution of Quality Metrics during Training with LRF}
\end{figure}

In \autoref{lrfcifar10} and \autoref{lrfcifar100} we observe high noise perturbation at initialization that quickly fades in the first 10 epochs, then rises subtly and plateaus at a low value. The residual noise extracted by LRF fades, as we'd expect of the random noise from initialization; it is replaced with learned structure, though it remains significant at later epochs with CIFAR100. All raw measures fail to mirror test accuracy and generalization gap at early epochs. Stable quality and effective rank measures begin mirroring test accuracy and generalization gap at around 10 epochs, and Frobenius and spectral norm measures only begin at the $60$th epoch. The quality metrics do a better job of mirroring test accuracy and generalization gap at earlier epochs with LRF. Frobenius and spectral norm measures, however, still follow a distinct drop from epoch 10 to 60 with LRF. It intuitively follows that we can expect stable quality and effective rank measures to correlate better with test accuracy and generalization gap, notably with LRF preprocessing, through all stages of training.

\section{Results Demonstration}

For a comprehensive display of experimental results please refer to the Supplementary Material.

\subsection{Generating a Family of Trained Models}
To test the effectiveness of the measures for tracking generalization performance at earlier stages of training, we train families of models with varied hyperparameter and channel size configurations, then save model weights and performances at each epoch. Models are trained for 70 epochs on CIFAR10 and CIFAR100 with various optimizers. We dub this dataset Generalization Dataset for Probeable Measures (GenProb).

\begin{table}[!b]
\caption{GenProb Model Architecture}
\label{gendamodelarchitecture}
\begin{center}
\renewcommand{\tabcolsep}{1pt}
\renewcommand{\arraystretch}{1.0}
\scriptsize{
\begin{tabular}{c||c||c}
    \hlinewd{1pt}
    \textbf{Block Index} & \textbf{Block Type} & \textbf{Output Shape} \\ \hline\hline
    0 & input & 32 x 32 x 3\\\hline
    1 & 3 x 3 convolution & 32 x 32 x 8\\\hline
    2 & convolutional block & 32 x 32 x \{40, 48\}  \\\hline
    3 & residual block & 18 x 18 x \{40, 48\}  \\\hline
    4 & convolutional block & 18 x 18 x \{40, 48\} \\\hline
    5 & residual block & 9 x 9 x \{40, 48\} \\\hline
    6 & convolutional block & 9 x 9 x \{40, 48\}   \\\hline  
    7 & global average pooling & 1 x 1 x \{40, 48\} \\\hline
    8 & linear &  1 x 1 x \{10, 100\} \\\hline
    \hlinewd{1pt}
\end{tabular}
}
\end{center}
\end{table}

\begin{figure}[!t]
    \includegraphics[width=0.38\textwidth]{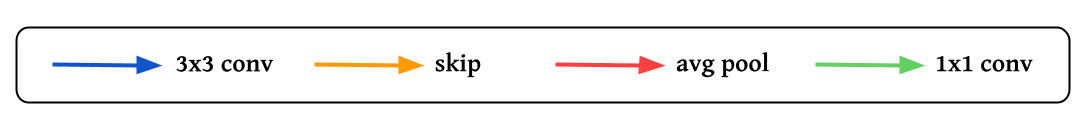}
    \centering
    \subfigure[Convolutional Block]{
    \label{convblockarch}
    \includegraphics[width=0.20\textwidth]{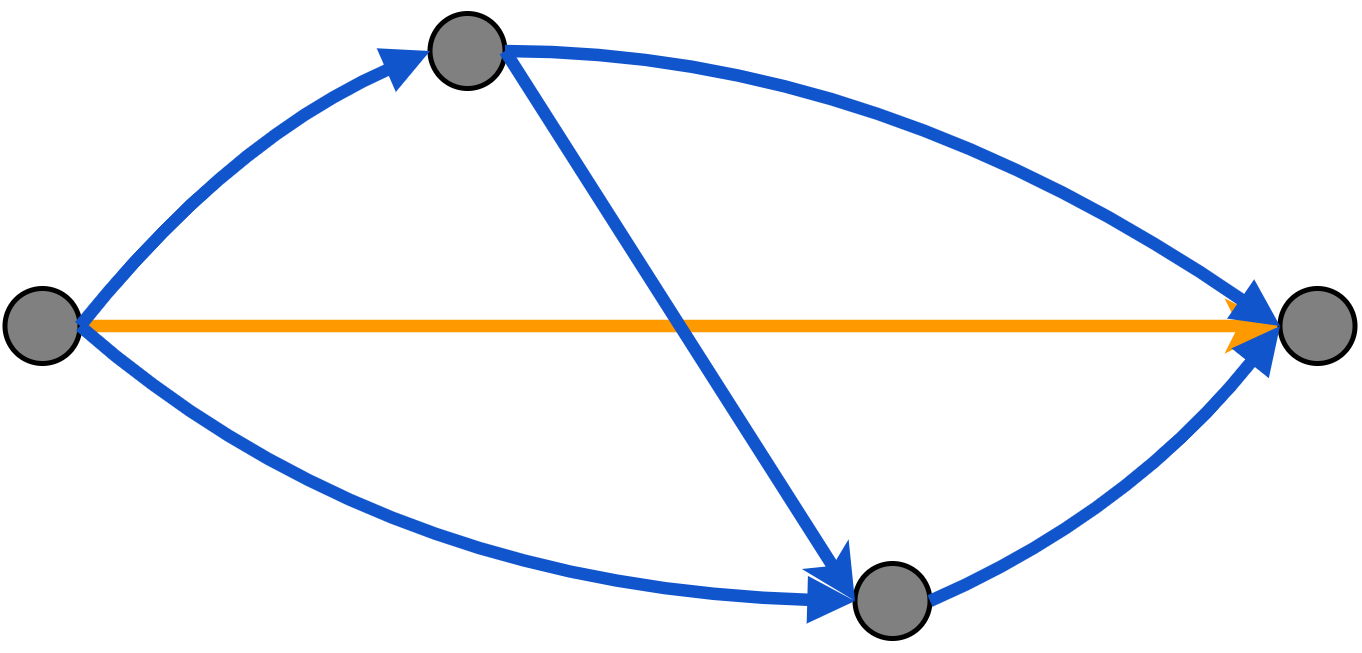}}
    \subfigure[Residual Block]{
    \label{resblockarch}
    \includegraphics[width=0.20\textwidth]{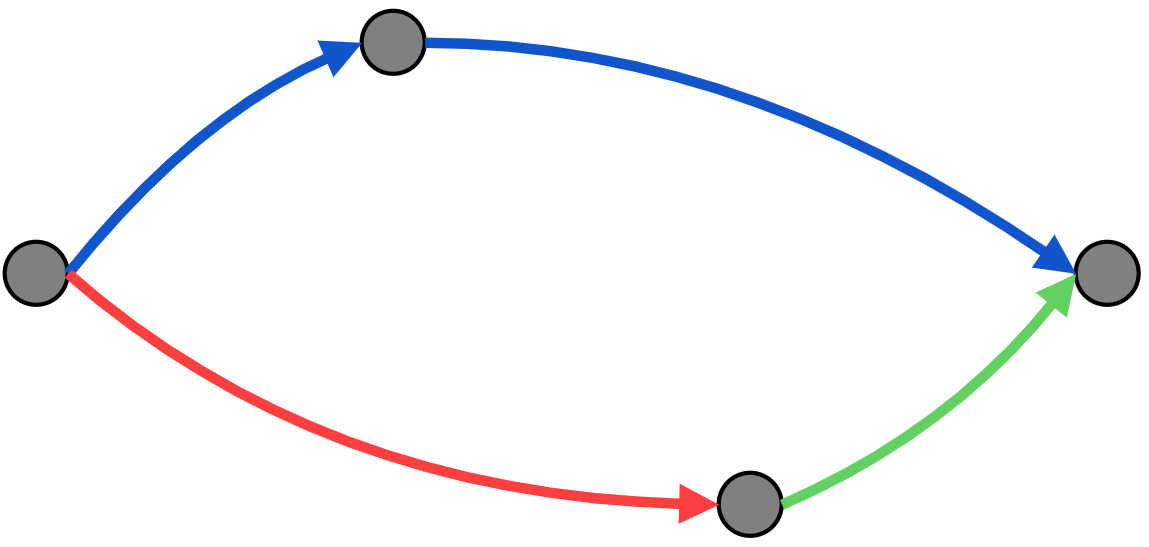}}
    \caption{GenProb Model Block Architecture}
    \label{blockarchs}
\end{figure}

The model architecture used is the same as that in NATS-Bench's size search space \cite{Dong_2021}, described in \autoref{gendamodelarchitecture}. The convolutional blocks can be described as directed acyclic graphs with five nodes of activations, as depicted in \autoref{convblockarch}. All nodes are ordered, and each node is connected to all nodes in front of it with a 3x3 convolution. The longest connection in the convolutional block (first node to last node) is replaced by a skip connection. The residual block is composed of a main path and a shortcut path, as illustrated in \autoref{resblockarch}. The main path contains a 3x3 convolution (stride 2) followed by another 3x3 convolution (stride 1), and an average pooling layer (stride 2) followed by a 1x1 convolution in the shortcut path. All 3x3 convolutions are followed by batch normalizations, and all changes in channel sizes take place at the earliest layers possible in the blocks. All activation functions are ReLU, including the prediction layer. The output channel sizes of blocks 2-6 are varied, the output channel size of block 7 depends on block 6, and the output channel size of block 8 depends on the dataset. 

The family of trained models is generated by varying initial learning rate, weight decay and channel sizes with the options listed in \autoref{modelfamilyhps}. Furthermore, input data is normalized channel-wise and augmented by random 32 x 32 crops with 4 pixels of padding and random horizontal flips of probability 0.5. Five repeated blocks in the model have their output channel sizes chosen, totaling $2^5 = 32$ permutations. From the set of all choices of hyperparameters we generate $5 \times 3 \times 32 = 480$ combinations, each yielding a unique trained model. We repeat with 7 different optimizers for both CIFAR10 and CIFAR100, training $480 \times 7 \times 2 = 6720$ unique configurations. Saving the model weights and performance at every epoch, we generate a total of $6720 \times 70 = 470400$ unique trained model files for our dataset.\footnote{GenProb will be released upon publication, alongside the GitHub repository.}
\begin{figure*}[t]
    \begin{center}
    \subfigure[Scatter Plots of Test Accuracy over \(Q_{E}^{L2}\) for Adam on CIFAR10]{
    \includegraphics[width=0.185\textwidth]{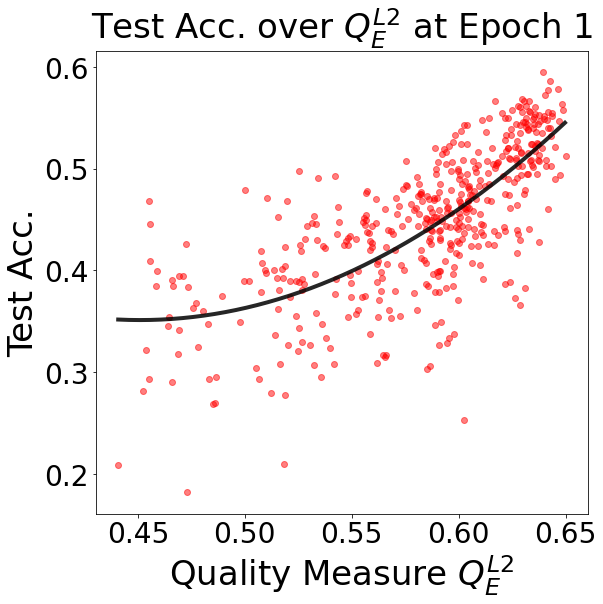}
    \includegraphics[width=0.19\textwidth]{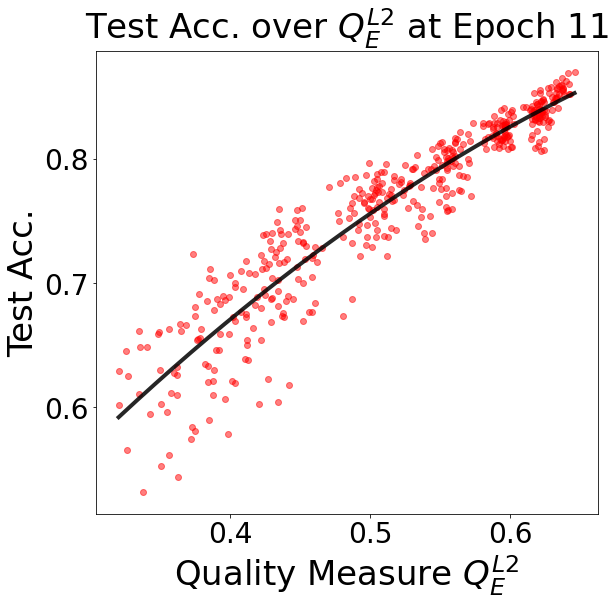}
    \includegraphics[width=0.19\textwidth]{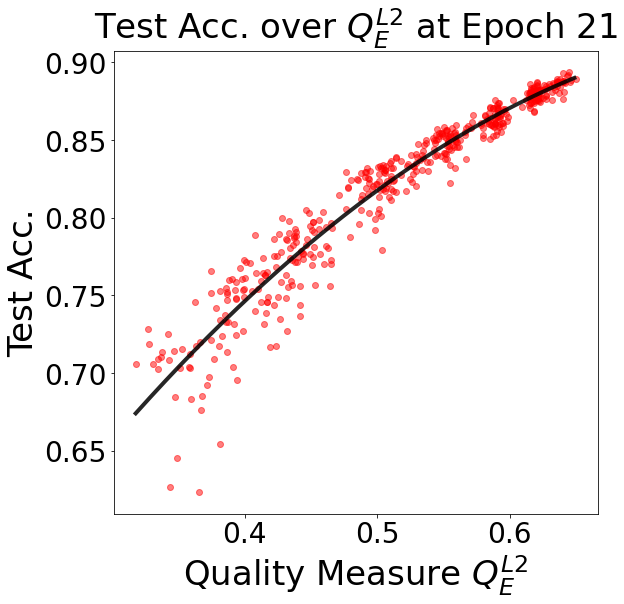}
    \includegraphics[width=0.19\textwidth]{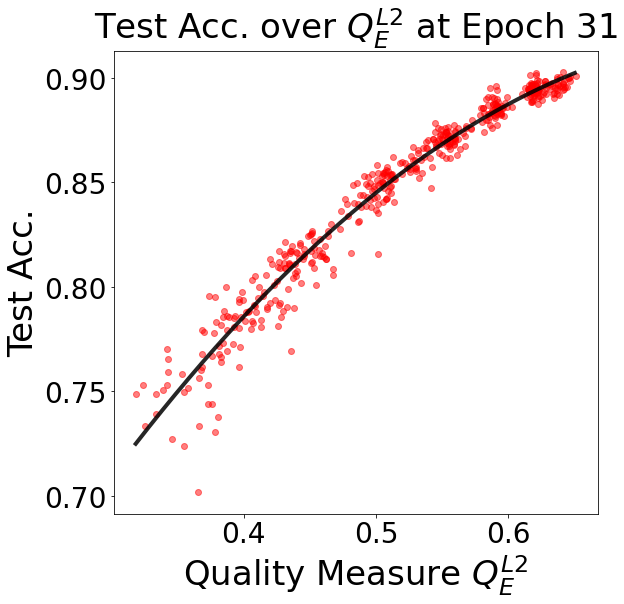}
    \includegraphics[width=0.19\textwidth]{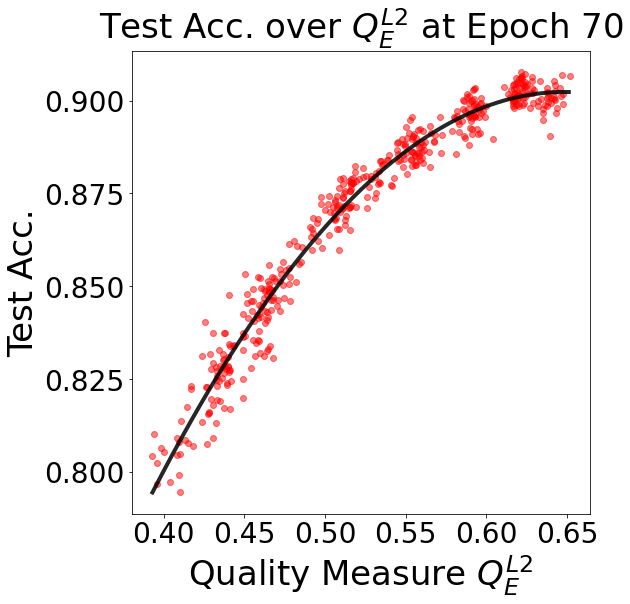}
    \label{AdaMTEST}
    }
    \subfigure[Scatter Plots of Generalization Gap over \(Q_{E}^{L2}\) for Adam on CIFAR10]{
    \includegraphics[width=0.185\textwidth]{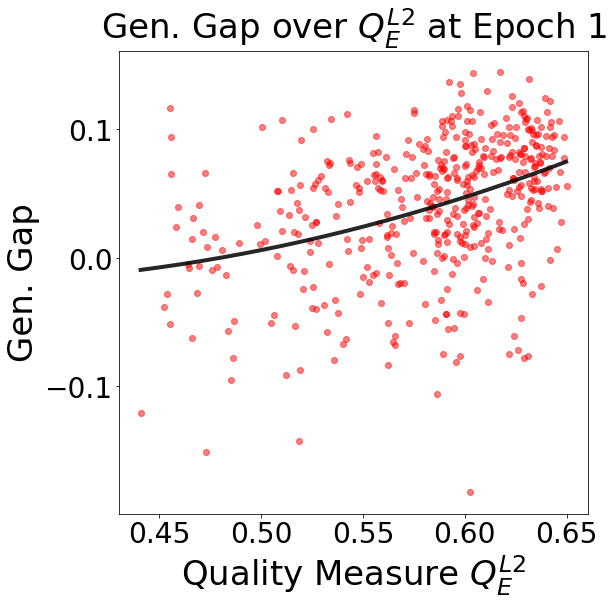}
    \includegraphics[width=0.19\textwidth]{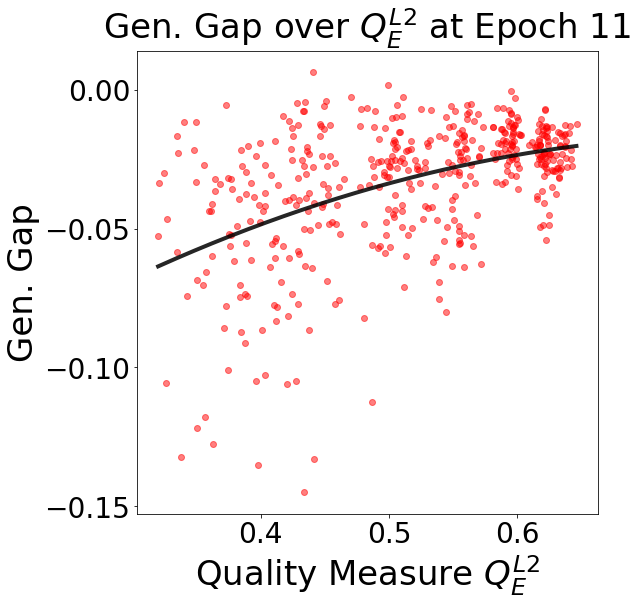}
    \includegraphics[width=0.19\textwidth]{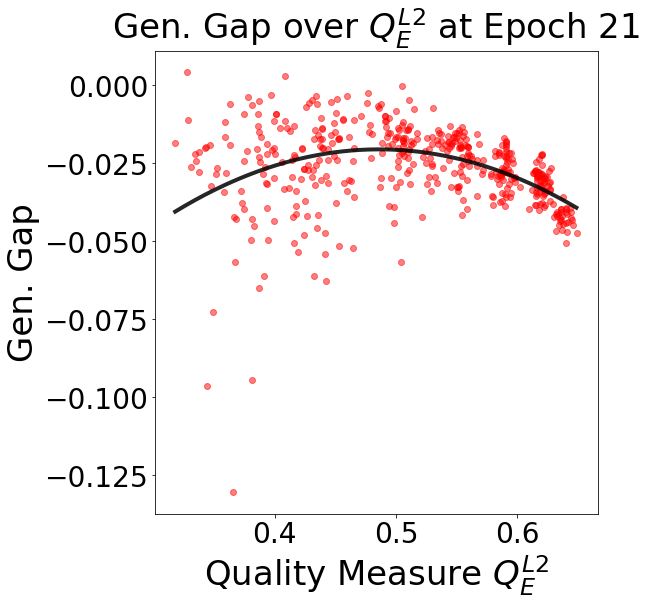}
    \includegraphics[width=0.19\textwidth]{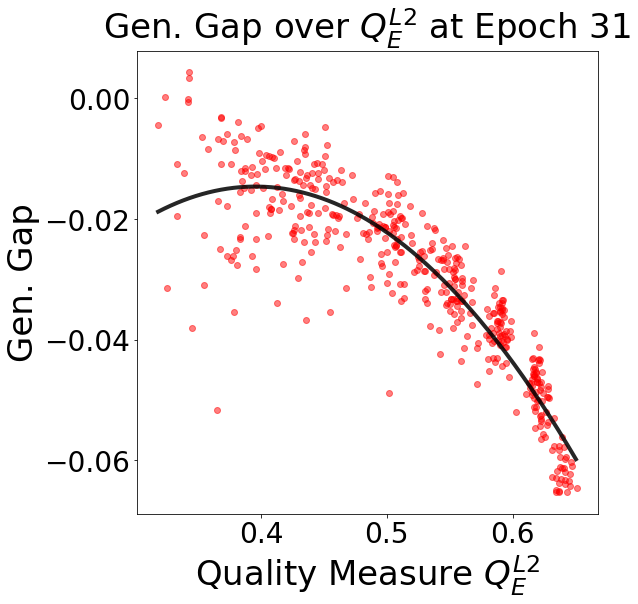}
    \includegraphics[width=0.19\textwidth]{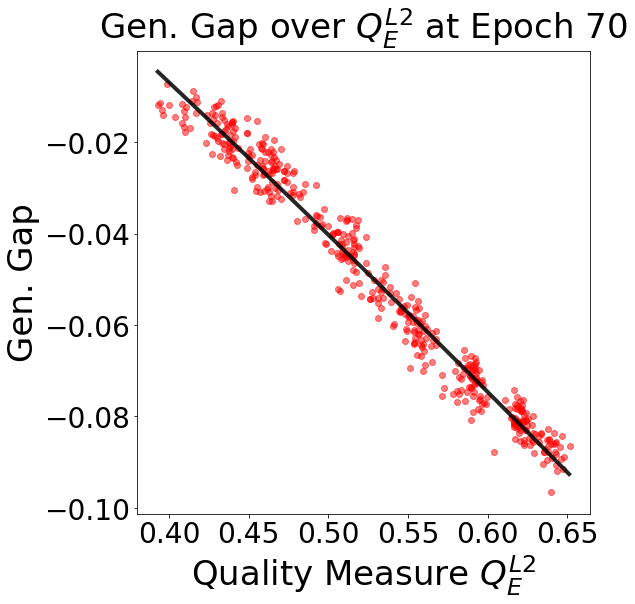}
    \label{AdaMGAP}
    }
    \includegraphics[width=0.27\textwidth]{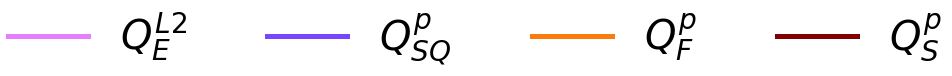}\hspace{95pt}\includegraphics[width=0.27\textwidth]{images/Correlation_Maps/legend.png}
    \subfigure[Quality Measure Correlations for Adam on CIFAR10]{
    \includegraphics[width=0.222\textwidth]{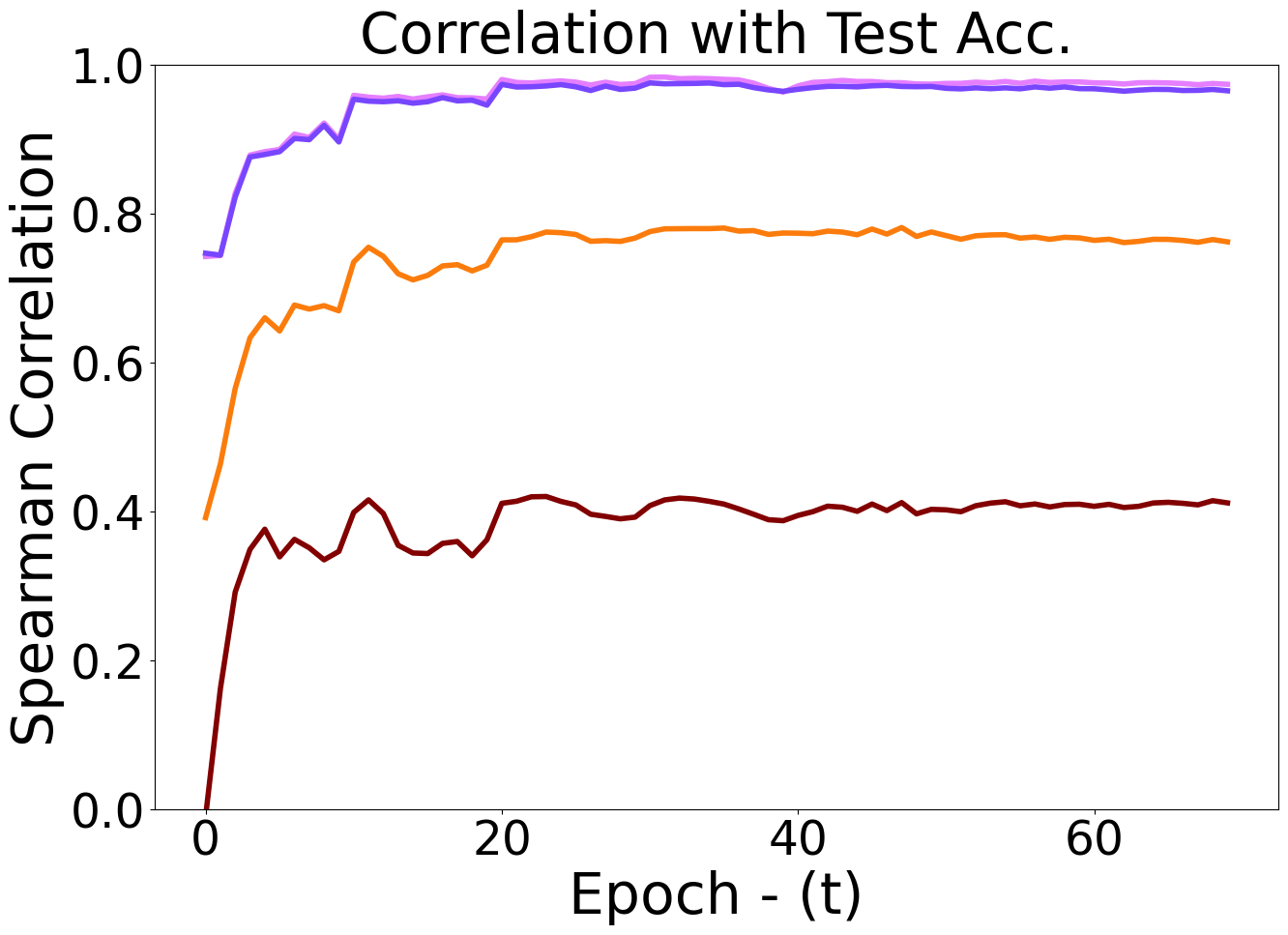}
    \includegraphics[width=0.23\textwidth]{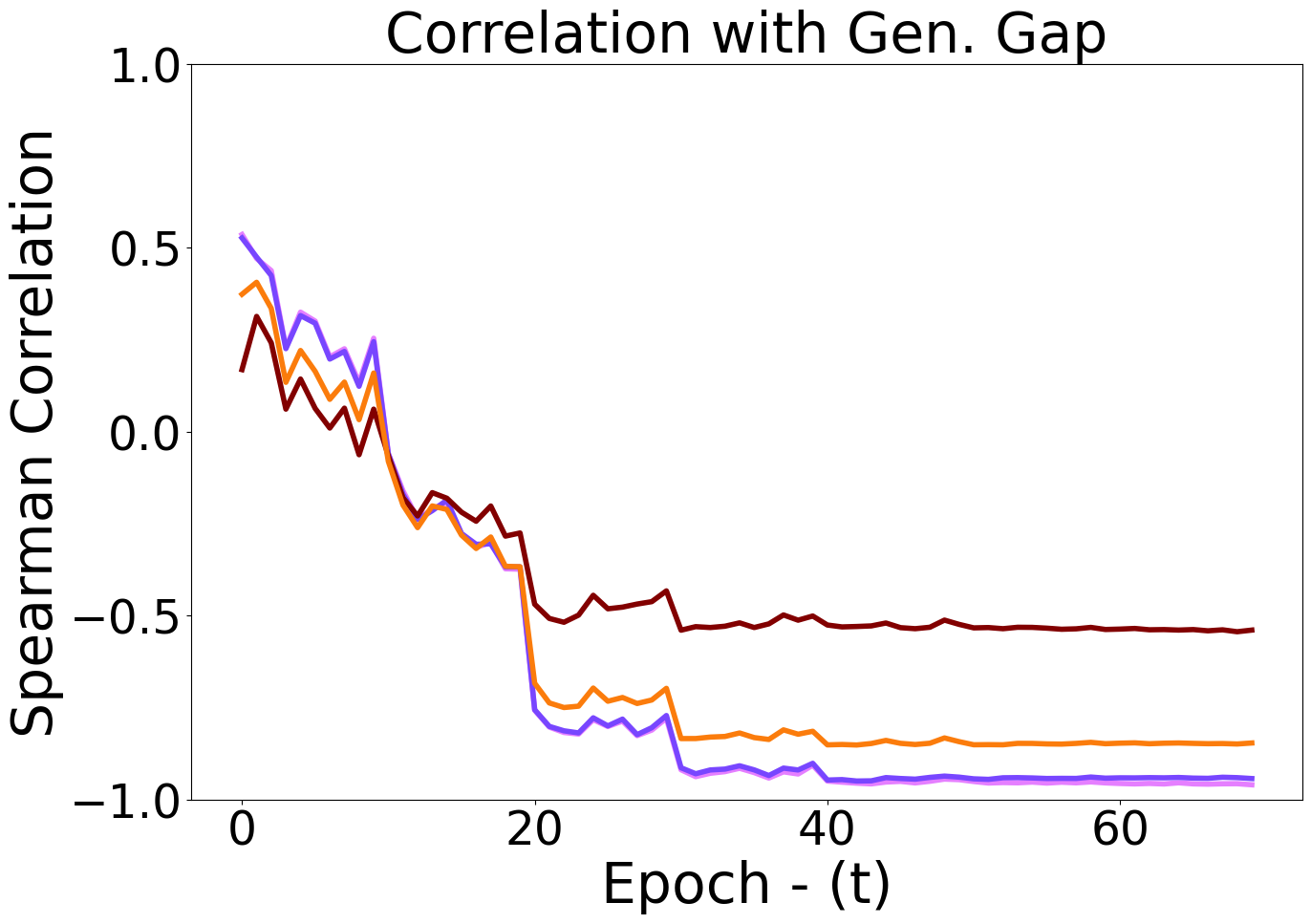}
    \label{corradamcifar10}
    }\hspace{10pt}
    \subfigure[Quality Measure Correlations for Adam on CIFAR100]{
    \includegraphics[width=0.222\textwidth]{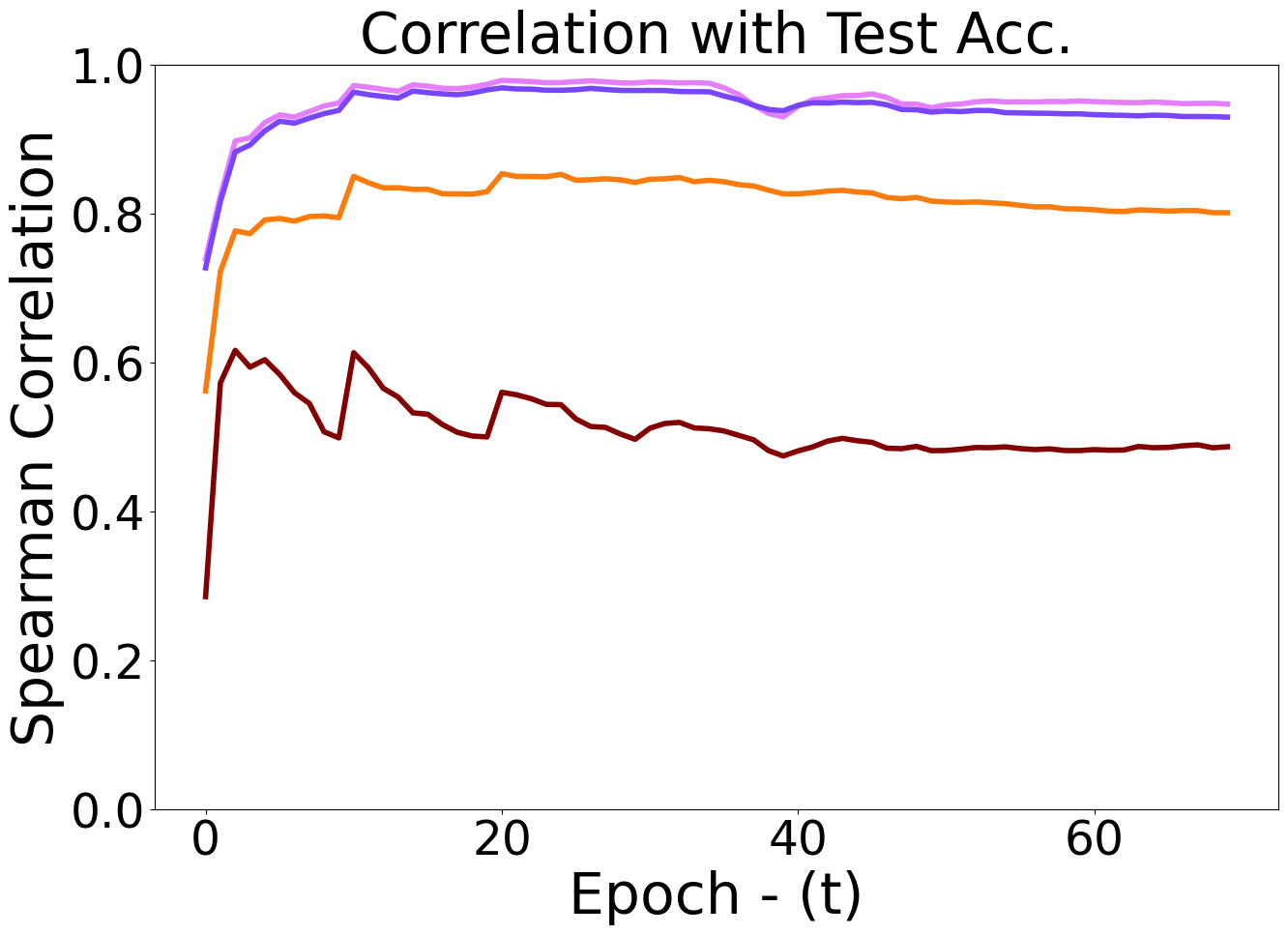}
    \includegraphics[width=0.23\textwidth]{images/Correlation_Maps/LilJon_AdaM_CIFAR100Correlation_with_Gen_Gap.png}
    \label{corradamcifar100}
    }
    \label{GENDAPRO}
    \caption{Testing Quality Measures with GenProb}
    \end{center}
\end{figure*}
\subsection{Effective Rank, Test Accuracy and Generalization Gap}
To visualize the relationship between the quality metrics and both generalization gap and test accuracy, we produce scatter plots of test accuracy and generalization gap over the quality metrics. Furthermore, by organizing these separate scatter plots relative to the quantity of training each set of models has undergone, we can study the evolution of the relationship. The quality metrics bias at earlier epochs due to residual noise from random initialization; we observe a lack of form in the effective rank scatter plots on models trained with Adam on CIFAR10 in \autoref{AdaMTEST} and \autoref{AdaMGAP} at earlier epochs. The quality metric evolves into clear, strong trends with test accuracy and generalization gap as training progresses and learned structure develops in the model weights. We observe similar behaviour with all three other metrics provided in the Supplementary Material (Fig. 1-16), and LRF only seems to weaken any trends, see Supplementary Material (Fig. 17-32).

\begin{table}[!b]
\caption{GenProb Hyperparameter Variation Summary}
\label{modelfamilyhps}
\begin{center}
\renewcommand{\tabcolsep}{1pt}
\renewcommand{\arraystretch}{1.0}
\scriptsize{
\begin{tabular}{c||c|c|c|c|c|c|c}
    \hlinewd{1pt}
    \textbf{Hyperparameter} & \textbf{RMSGD} & \textbf{SGD} & \textbf{SAM} & \textbf{SGDP} & \textbf{Adam} & \textbf{AdaBound} & \textbf{AdamP}  \\ \hline\hline
    Learning Rate & \multicolumn{4}{c|}{0.02, 0.04, 0.06, 0.08, 0.10} & \multicolumn{3}{c}{2e-3, 4e-3, 6e-3, 8e-3, 0.01} \\\hline
    Weight Decay  & \multicolumn{7}{c}{1e-4, 5e-4, 1e-3}  \\\hline
    Channel Sizes & \multicolumn{7}{c}{40, 48}            \\\hline
    Momentum      & \multicolumn{7}{c}{0.9}             \\\hline
    Batch Size    & \multicolumn{7}{c}{256}               \\\hline
    Epochs        & \multicolumn{7}{c}{70}                \\\hline
    Scheduler Step Size & - & \multicolumn{6}{c}{10}     \\
    \hlinewd{1pt}
\end{tabular}
}
\end{center}
\end{table}

We observe a clear linear relationship between the effective rank measure and generalization gap at later epochs, and a 2nd order relationship between the effective rank measure and test accuracy. The plateauing trend with test accuracy delineates a bound on test accuracy; maximizing effective rank above this bound would still increase generalization gap (linear trend) however, suggesting an increase in train accuracy without changes in test accuracy. It is still evident that for a model trained on CIFAR10 with Adam, a greater effective rank indicates greater test accuracy, and a greater (negative) generalization gap. In fact, the relationships take form before model completion, as trends with test accuracy show as early as epoch 10, and trends with generalization gap show at epoch 40; this may be of interest for implementation of effective rank in NAS, HPO and optimization algorithms. We find similar timelines with all other quality metrics, see Supplementary Material (Fig. 1-32).

\subsection{Quality Measure Correlation Evolution} 
By plotting the correlations of the quality metrics with test accuracy and generalization gap in \autoref{corradamcifar10} and \autoref{corradamcifar100}, we can understand the relative progression of the effectiveness of these measures through different stages of training. As the aforementioned trends in \autoref{AdaMTEST} and \autoref{AdaMGAP} become more distinct, the corresponding correlations increase in magnitudes, some nearly up to 1. LRF preprocessing of weights only decreases the magnitude of correlation through all stages of training, see Supplementary Material (Fig. 41-42)

The large correlations indicate robustness to changes in training hyperparameters, and model channel sizes. Effective rank and stable quality measures prove to be the most effective and robust generalization measures through all training phases and across dataset complexities. The Frobenius and spectral norm measures underperform consistently and are less consistent across dataset complexities, yielding lower correlations with CIFAR10. Also, as previously highlighted in the scatter plots, correlations with test accuracy plateau after 10 epochs and correlations with generalization gap plateau after 40 epochs.

\section{Conclusion}
In this work we investigated "explainable" generalization measures that can probe individual layers of a deep neural network. We tested four quality metrics, that are calculated from layer weight tensors, over spaces of similar model architectures for NAS (NATS-Bench). We analyzed the evolution of the quality metrics during training, then produced a dataset for more meaningful analysis and testing related to HPO and training optimization. We introduced GenProb, a dataset of 470,400 trained models and their performance metrics, distinguished by hyperparameter and channel size variations. We demonstrated effective rank and stable quality measure effectiveness and robustness to variations in training hyperparameters, channels sizes, dataset complexities and stages of training. Furthermore, we investigate the amount of training required to produce meaningful generalization measures from model weights, and the shape of the relationships of these measures with test accuracy and generalization gap.

We hope this work inspires and guides the production of novel NAS, HPO and training optimization algorithms that leverage probeable generalization measures to maximize model generalization performance at a layer level. We also aim to motivate further development of probeable generalization measures, for which GenProb will prove a useful tool. Our investigated measures don't prove robust to drastic variations in model architecture, and as such may not be suitable for all NAS algorithms. We also only consider simple convolutional neural networks and two datasets (CIFAR10 and CIFAR100); we hope that future investigations will experiment with other types of model, deeper models, and other datasets.

\bibliographystyle{IEEEtran}
\bibliography{QC-Explaining-Generalization}
\end{document}